\documentclass[11pt,a4paper]{article}
\usepackage[hyperref]{acl2021}
\usepackage{times}
\usepackage{multirow}
\usepackage{latexsym}

\usepackage[utf8]{inputenc}
\usepackage{graphicx}
\usepackage{subfigure}
\usepackage{enumitem}
\usepackage{xcolor}
\usepackage[ruled,vlined]{algorithm2e}
\usepackage{microtype}
\usepackage{amsmath}
\mathchardef\mhyph="2D

\aclfinalcopy 
\def\aclpaperid{2234} 

\title{ZmBART: An Unsupervised Cross-lingual Transfer Framework for Language Generation}

\author{\textbf{Kaushal Kumar Maurya} \\
  Indian Institute of Technology Hyderabad \\
  Hyderabad, India \\
  \texttt{cs18resch11003@iith.ac.in} \\ \\
  \textbf{Yoshinobu Kano} \\
  Shizuoka University, Japan \\
  \texttt{kano@inf.shizuoka.ac.jp} 
   \And
  \textbf{Maunendra Sankar Desarkar} \\
  \hspace{10mm} Indian Institute of Technology Hyderabad \\
  Hyderabad, India \\
  \texttt{maunendra@cse.iith.ac.in}   \\ \\
  \textbf{Kumari Deepshikha} \\
  NVIDIA, India \\
  \texttt{deepkshikha@gmail.com}}
  
\date{}

\begin{document}

\maketitle

\begin{abstract}
Despite the recent advancement in NLP research, cross-lingual transfer for natural language generation is relatively understudied. In this work, we transfer supervision from high resource language (HRL) to multiple low-resource languages (LRLs) for natural language generation (NLG). We consider four NLG tasks (text summarization, question generation, news headline generation, and distractor generation) and three syntactically diverse languages, i.e., English, Hindi, and Japanese. We propose an unsupervised cross-lingual language generation framework (called ZmBART) that does not use any parallel or pseudo-parallel/back-translated data. In this framework, we \textit{further} pre-train mBART sequence-to-sequence denoising auto-encoder model with an auxiliary task using monolingual data of three languages. The objective function of the auxiliary task is close to the target tasks which enriches the multi-lingual latent representation of mBART and provides good initialization for target tasks. Then, this model is fine-tuned with task-specific supervised English data and directly evaluated with low-resource languages in the Zero-shot setting. To overcome catastrophic forgetting and spurious correlation issues, we applied freezing model component and data argumentation approaches respectively. This simple modeling approach gave us promising results. We experimented with few-shot training (with 1000 supervised data-points) which boosted the model performance further. We performed several ablations and cross-lingual transferability analysis to demonstrate the robustness of ZmBART. 
\end{abstract}

\section{Introduction}
\label{sec:intro}

\label{subsec:few_sup}
\begin{figure}[!htb]
    \centering
    \includegraphics[width=7.5cm]{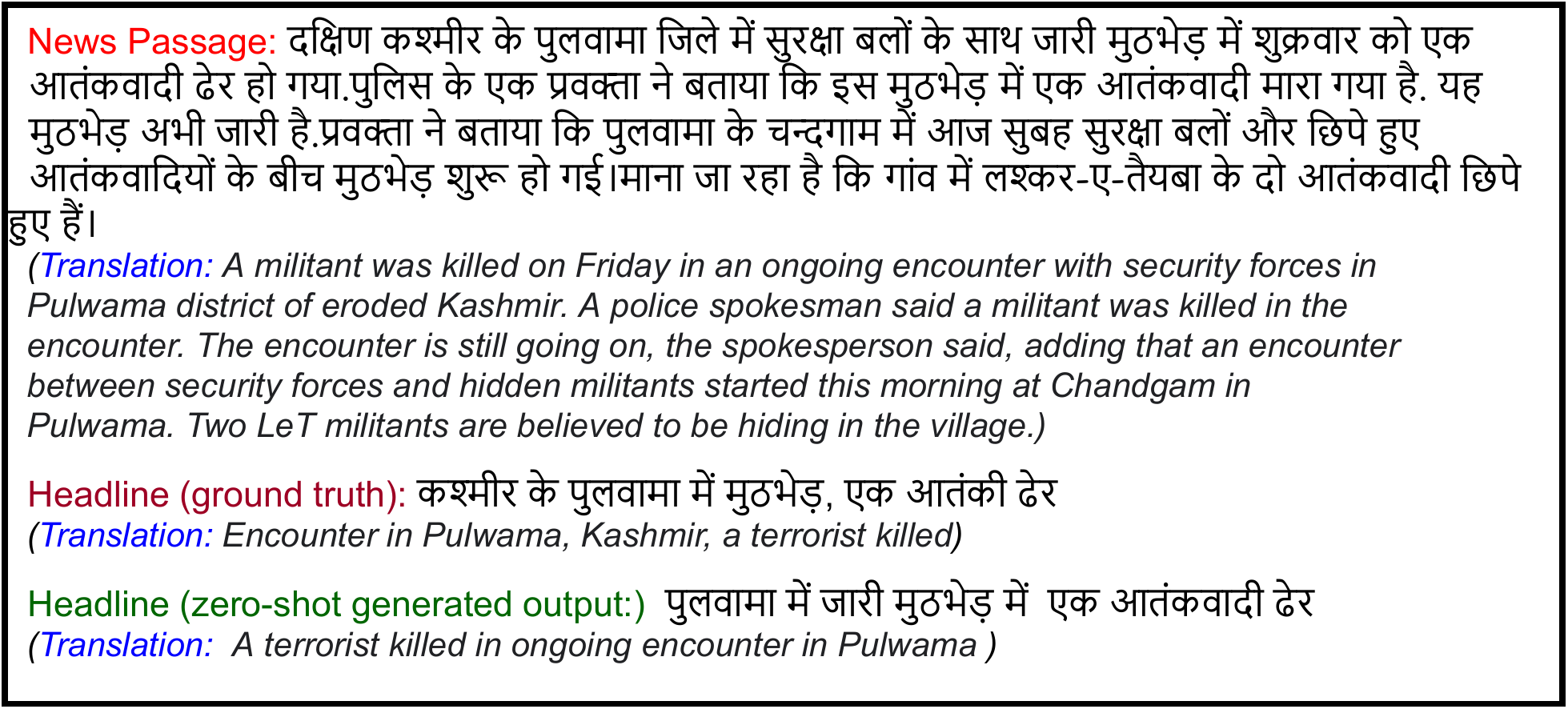}
    \caption{Zero-shot news headline generation from ZmBART in Hindi language}
    \label{fig:sample_data}
    \vspace{-.1in}
\end{figure}

Recent advancement in natural language generation (NLG) is heavily oriented towards large annotated training data. Such large task-specific annotated data is available for high resource language (HRL) like English. The tasks become challenging when limited training data is available. This is often observed for low-resource languages (LRLs) like Hindi, Japanese, etc. Manually annotating large data is time-consuming, expensive and uninteresting. This limits the model development and product deployment for LRLs. Moreover, despite large active research in cross-lingual representation learning \cite{hu2020xtreme, conneauetal2020unsupervised, liangetal2020xglue}, the area of cross-lingual transfer and generation is relatively under-explored. Motivated by these factors, we propose a novel framework to transfer supervision from HRL to LRLs where model is trained on one language and directly evaluated for unseen languages. This enables cross-lingual transfer and generation for low resource languages in zero and few-shot settings for different tasks. The framework can be easily extended to other tasks and languages.

We carefully selected four challenging NLG tasks i.e., news headline-generation (NHG), question generation (QG), abstractive text summarization (ATS) and distractor generation (DG) to validate the framework's performance. NHG and ATS require understanding of input passage to generate meaningful headline and summary respectively. QG task should accumulate information from a passage and answer to generate high-quality questions. Distractor generation is the task of generating incorrect options from reading comprehension MCQ. It is challenging because generated distractors should be in the context with question but should not be semantically equivalent to the answer. We consider two LRLs i.e., Hindi and Japanese from two different language families. English is selected as the HRL from which the learning would be transferred to the LRLs. 
All three selected languages are different in their syntactic structures and typologically diverse. As there is no established publicly available dataset for DG in Hindi, we also create a new DG dataset for Hindi called as \textbf{HiDG}\footnote{HiDG dataset download link: \url{https://github.com/kaushal0494/ZmBART}}.

Our proposed framework to achieve this transfer of supervision from HRL to LRL under multiple languages and multiple tasks is named as ZmBART. ZmBART is based on mBART \cite{liu2020multilingual}, a pre-trained model for cross-lingual natural language generation (NLG).  
We \textit{further} pre-train mBART with a novel auxiliary task. Then the trained model is fine-tuned on large task-specific supervised data in English and evaluated directly on Hindi and Japanese languages in zero/few-shot setting for the tasks under consideration. We observe that the auxiliary task plays a critical role on the model's performance and needs to be carefully designed. 
This framework can be directly applied to multiple cross-lingual generation tasks without even the need to modify model hyper-parameters.  
Figure-\ref{fig:sample_data} shows a zero-shot NHG sample output generated by the ZmBART model. 
Our main contributions in this work can be summarized as:
\begin{enumerate}
    \itemsep0em
    \item We propose a novel zero-shot cross-lingual generation framework called ZmBART without parallel data and without back-translation. The framework can be directly applied across multiple tasks without even modifications in hyper-parameter values. 
    \item We demonstrate the effectiveness of ZmBART on four cross-lingual generation tasks across three typologically diverse languages. 
    \item We have created \textit{HiDG}, a high-quality distractor generation dataset for the Hindi language.
\end{enumerate}

\section{Related Work}
\label{sec:rel_work}
Early works on cross-lingual generation rely on machine translation (MT). In the very first work, \citet{wanetal2010cross} leveraged the MT pipeline for cross-language document summarization. They first translate the non-English test instances to English. This translated text is fed through the supervised model (trained with document summarization data in English) to generate English summaries. Finally, these summaries are translated back to the target language. \citet{ 10.1109/TASLP.2018.2842432} and \citet{duanetal2019zero} used MT systems to generate pseudo training data for cross-lingual summarization and news headline generation respectively. However these MT based models are not suitable for low resource languages as they do not share parameters across-languages and generated translations are error-prone. 

Recently there are a few works in the direction of supervision transfer from HRL(s) to LRL(s) for language generation. \citet{kumaretal2019cross} used back-translation (needs MT system) and annotated supervised data for cross-lingual question generation. \citet{Chi_Dong_Wei_Wang_Mao_Huang_2020} used parallel data to train a sequence-to-sequence model for zero-shot cross-lingual abstractive text summarization and question generation. \citet{DBLP:journals/corr/abs-2006-15020} proposed a pre-training based on mono-lingual paragraphs. Then this pre-trained model is used for zero-shot abstractive text summarization (ATS) in multiple languages. They trained a model on the ATS dataset on all the languages except the test language. This approach needs annotated data in multiple languages. Existing supervision transfer methods require parallel data for the cross-lingual tasks. Either they use available parallel corpora directly, or they translate/ back-translate data to generate pseudo-parallel corpora. Both these approaches pose significant challenges, as task-specific parallel data for multiple languages is difficult to obtain, and MT are far from perfect, especially for low resource languages. 

Unlike the previous approaches, we did not use any parallel data or back-translation in our proposed framework. We did not pre-train any model from scratch. Instead, we leveraged the existing pre-trained model mBART. We included four challenging generation tasks across three syntactically diverse languages. Even we did not modify any hyper-parameters across the tasks and languages. 
All these considerations make the framework simple and easy to use. Further, it enables the addition of different other languages and NLG tasks in the proposed framework a simple extension exercise.

\section{Methodology}
\label{sec:methodlgy}
\begin{figure}[h]
    \centering
    \includegraphics[width=7cm]{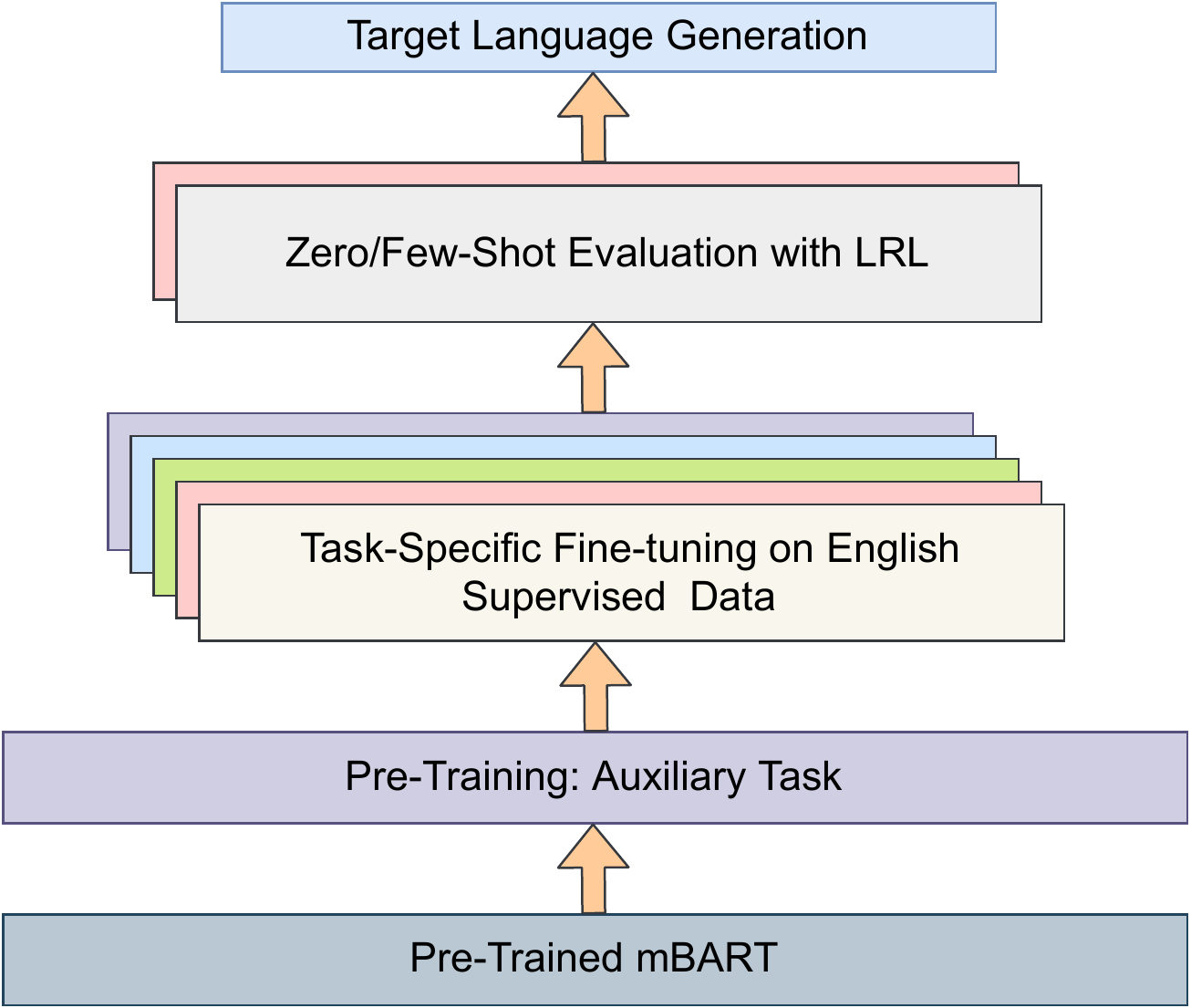}
    \caption{Architecture diagram of ZmBART}
    \label{fig:arct_digm}
    \vspace{-.1in}
\end{figure}

Figure \ref{fig:arct_digm} shows an outline of our proposed ZmBART framework. ZmBART is based on pre-trained mBART \cite{liu2020multilingual} model. In our framework, we take the mBART model and \textit{further pre-train} it on an auxiliary task. The auxiliary task is designed in such a way that the objective function of auxiliary task is close to fine-tuning tasks and only utilizes the mono-lingual data from the selected languages. 
Similar to mBART model we use language identifier tag with slight modification. We concatenate $<fxx><2xx>$ tags in input data instance where $xx$ indicates the language tag. Given an input sentence and the language tag the model encodes the sentence in multi-lingual space. By conditioning on the encoded representation and language tag the decoder generates output text in target language.

\subsection{Multilingual BART (mBART)}
\label{subsec:mbart}
Multilingual BART \cite{liu2020multilingual} is an extension of BART model \cite{lewis-etal-2020-bart} to multiple languages. It is a transformer-based sequence-to-sequence pre-trained model. The model is trained on monolingual data in many languages from Wikipedia Common Crawl corpus with BART language model objective. Particularly, The training data is concatenation of data from $K$ languages i.e., $\mathcal{D}=\{\mathcal{D}_1, \mathcal{D}_2 \dots \mathcal{D}_K\}$ where $\mathcal{D}_i$ is a collection of monolingual documents in language $i$. They introduced two types of noises to corrupt the text: (1) random token span masking and (2) sentence order permutation. mBART is trained as denoising autoencoder. During training, the model has to predict text $X$ from it's corrupted version $g(X)$, where $g$ is noise function. The aim is to maximize the following objective function
\begin{equation}
    \mathcal{L}_\theta = \sum_{\mathcal{D}_i \in \mathcal{D}} \sum_{x \in \mathcal{D}_i} log P(x|g(x);\theta),
\end{equation}
where $x$ is a data instance of language $i$. Probability distribution $P$ is defined by the sequence-to-sequence model. mBART gave state-of-the-art results in sentence and document level machine translations tasks. 
Details about mBART model can be found in \citet{liu2020multilingual}.       

\subsection{Unsupervised Auxiliary Task}
\label{subsec:auxi}
Although the mBART pre-trained model encodes a multi-lingual latent space, it can not be used directly for cross-lingual generation. This is because the model is jointly trained on denoising objectives which do not directly follow auto-regressive decoding, thereby causing mismatch between pre-training and fine-tuning objectives. To overcome this problem, an unsupervised auxiliary task is introduced. We design the auxiliary task with the following desiderata in mind. It (1) should only utilize mono-lingual data from selected languages, (2) should enrich the mBART latent representations for selected languages and (3) train the decoder in pure auto-regressive manner with a training objective which is close to multiple fine-tuning tasks.

The auxiliary task in ZmBART is an additional pre-training step for \textit{better warm-start} to downstream auto-regressive NLG tasks - although the final task (Distractor/Question/Summary generation) can be different from the auxiliary task. Additionally, this step allows the model to have a closer look at the languages under consideration and enrich/adjust the representations and parameters accordingly.

Outputs of the NLG tasks considered in this work are expected to contain words from different parts of the input. Generation of the output tokens are handled by the framework using an encoder-decoder setup. Hence we decide to have an auxiliary task that also encodes the input, and attends to this encoded representation to generate the output words in auto-regressive manner. This way, a single auxiliary task can help to enrich the token representations, warm up the encoder-decoder weights for fine tuning, and also caters to the multiple final output tasks. We define the auxiliary task as: \textit{Given an input passage, generate few random sentences (called rand-summary) from the passage}. After experimentation we found that randomly generating 20\% sentences from passage works the best. Particularly, the input passage has length between $5 \mhyph 25$ sentences and output is $1 \mhyph 5$ random sentences from the passage. We do not assume any relations among sentences of the passage. We sample equal proportion of monolingual data from three languages. Data preparation steps for the auxiliary task are given below:
\vspace{-.05in}
\begin{enumerate}
    \itemsep-0.5em
    \item Generate a random number $k \in \{5. \cdots, 25\}$. $k$ denotes the size of input passage
    \item \textsc{passage:} Append $k$ continuous sentences, starting from a random index of monolingual corpus $D_i$ of the $i^{th}$ language
    \item \textsc{rand-summary:} Randomly select 20\% sentences from the passage
    \item Repeat steps $1$ to $3$ for $p$ languages
    \item Repeat steps $1$ to $4$ for $N$ times, to collect $Np$ $<$\textsc{passage, rand-summary}$>$ pairs
\end{enumerate}



\subsection{Fine-Tuning on Downstream NLG Tasks}
\label{subsec:ftdstask}
The proposed pre-trained model is directly fine-tuned on four downstream tasks: Question Generation (QG), News Headline Generation (NHG),  Abstractive Text Summarization (ATS) and  Distractor Generation (DG). First, the model is fine-tuned on large task-specific English supervised data and then this trained model is directly evaluated on Hindi and Japanese evaluation datasets in zero-shot setting. To validate the hypothesis that the ZmBART framework is robust across multiple tasks and languages, we did not modify any hyper-parameters during fine-tuning. It is often observed that including a few instances from LRL to  supervised data boosts the model performance. To validate this point we further fine-tuned ZmBART with 1000 task-specific supervised data-points in Hindi and Japanese languages in few-shot setting which boosts the model performance.

\subsection{Dealing with Catastrophic Forgetting and Spurious Correlation}
\label{subsec:SolCF}
 During experimentation with the zero shot setup, it is observed that the model always generates the output text in English irrespective of input and language tag. We suspect this to be due to catastrophic forgetting problem \cite{van2019three}. The supervised training completely overrides/erases the pre-trained learning. The generator (decoder) becomes biased towards English due to the explicit supervision learned from large task-specific English data. To overcome this problem, we freeze all word embeddings and all the parameters of decoder layers during fine-tuning with English data. Although this resolves the problem for NHG, QG and DG, the problem did not get completely resolved for the ATS task. We noticed that the zero-shot ATS output now is not completely in English, but it became of code-mix nature. In other words, the number of English words in the output reduced, but still lot many English words remained. The code-mixed outputs were logical and meaningful. We assume this to be due to spurious correlation issue, also reported in \cite{guetal2019improved}. To resolve this issue, we added a few examples (25 in number) of the auxiliary-task data during the fine-tuning step. This augmentation was helpful to address the spurious correlation issue for ATS. It is to be noted that the non-English data used for this augmentation is still of unsupervised and monolingual nature.
\vspace{-2mm}

\section{Experimental Setup and Results}
\label{sec:expres}
We conduct experiments over four NLG tasks in three languages. We compare the performance of ZmBART  with strong and MT pipeline based baseline models. We use both automated and manual evaluation metrics to evaluate model performances. 
\vspace{-.3in}
\subsection{Baselines}
\label{subsec:base}
Prior results are not available in literature for selected languages and datasets. Hence, for performance comparison, we developed several strong baselines based on recent models and architectures. Details of these baselines are mentioned below:
\begin{itemize}[leftmargin=*,itemsep=-0.3em]
    \item \textbf{MT Pipeline (mBART):} Here, we fine-tune mBART on task-specific English data. Non-English test data instances are first translated into English and passed to the fine-tuned model. The output is translated back to the input language. Google Translator is used for translations.
    
    \item \textbf{mBART+\textsc{MADMo}:} This is an \textbf{mBART} based baseline where the auxiliary task has \textbf{M}asking \textbf{A}nd \textbf{D}enoising objective with \textbf{Mo}no-lingual data in three languages. The aim is to enrich the cross-lingual latent representation space of mBART for English, Hindi and Japanese. 
    \item \textbf{mBART+\textsc{MADPd}:} Inspired from  \cite{Chi_Dong_Wei_Wang_Mao_Huang_2020}, we took \textbf{P}arallel \textbf{D}ata (English-Hindi and English-Japanese) and concatenate each parallel instances of two languages. Then we used this data with  \textbf{M}asking \textbf{A}nd \textbf{D}oising objective to further train mBART. Including parallel data provides explicit supervision while generating Hindi and Japanese text.     
\end{itemize}

\subsection{Evaluation}
\label{subsec:eval}
We use both automated and manual evaluation metrics for performance comparison. Multiple metrics are used in literature for NLG tasks.  
Since we are considering multiple tasks, for brevity, against each task we only report values of the metrics commonly used by the community for that particular task. For automatic evaluation we used both lexical match (\textbf{BLEU} \cite{papinenietal2002bleu} and \textbf{ROUGE} \cite{lin2004rouge}) as well as embedding based evaluation metrics (\textbf{BERTScore} \cite{DBLP:conf/iclr/ZhangKWWA20}). To evaluate question generation and distractor generation tasks we use case-mix BLEU-4 (BL) score from  sacreBLEU implementation, ROUGE-L (R-L) and BERTScore (BS). For ATS and NHG tasks ROUGE-1, ROUGE-2  and ROUGE-L are used. 

We follow a similar approach for manual evaluation as \citet{Chi_Dong_Wei_Wang_Mao_Huang_2020}. We sampled 50 generated data points each for QG, ATS and NHG tasks in both Hindi and Japanese languages. We use three metrics: \textit{Fluency} (Flu), \textit{Relatedness} (Rel) and \textit{Correctness} (Corr). \textbf{Fluency} measures \textit{how fluent the generated text is}. \textbf{Relatedness} indicates \textit{how much the generated outputs are in the context with input(s)}, \textbf{Correctness} measures \textit{semantics and meaningfulness}. For DG, we use an additional metric called \textbf{Distractibility} that measures \textit{the degree of confusion for generated incorrect options}. For DG task, there can be large number of good distractors for given input, in such situation the manual evaluation is more reliable. We sample 100 generated outputs for DG task. We employed large pool of evaluators from native Hindi and Japanese speakers to evaluate Hindi and Japanese output texts respectively. We asked each annotator to rate the generated texts on a scale of 1-5 (1 is very bad and 5 is very good) for all the metrics. We intentionally selected outputs of ZmBART and two best baselines to reduce the evaluators workload.   


\begin{table*}
\centering
\scalebox{0.75}{
\begin{tabular}{l|c|c|c|c}
\hline\hline
\textbf{Model} & \textbf{News Headline Generation} & \textbf{Question Generation}& \textbf{Abstractive TS} & \textbf{Distractor Generation}\\
\hline
 \textbf{Metrics} & \textbf{R-1} \hspace{0.5cm} \textbf{R-2} \hspace{0.5cm}\textbf{ R-L} & \textbf{BL} \hspace{0.5cm} \textbf{R-L} \hspace{0.5cm} \textbf{BS} & \textbf{ R-1} \hspace{0.5cm} \textbf{R-2} \hspace{0.5cm} \textbf{R-L} & \textbf{BL} \hspace{0.5cm} \textbf{R-L} \hspace{0.5cm}\textbf{ BS} \\
\hline\hline
\multicolumn{5}{l}{\textit{Cross-lingual zero-shot generation results}} \\\hline
 MT Pipeline(mBART) & 16.61 \hspace{0.5cm} 4.91 \hspace{0.5cm} 15.83  & 2.6 \hspace{0.5cm} 21.31 \hspace{0.5cm} 71.53  & 11.15 \hspace{0.5cm} 3.11 \hspace{0.5cm} 10.93 & 1.6 \hspace{0.5cm} 9.66 \hspace{0.5cm} 67.35 \\
 
 mBART+\textsc{MADMo} & 29.32 \hspace{0.5cm} 16.36 \hspace{0.5cm} 27.52  & 3.9 \hspace{0.5cm} 23.70 \hspace{0.5cm} 73.76  & 18.25 \hspace{0.5cm} 4.92 \hspace{0.5cm} 16.10 & 2.8 \hspace{0.5cm} 15.86 \hspace{0.5cm} 72.26 \\
 
 mBART+\textsc{MADPd} & 24.02 \hspace{0.5cm} 13.41 \hspace{0.5cm} 23.29  & 4.3 \hspace{0.5cm} 25.29 \hspace{0.5cm} 73.74  & 10.47 \hspace{0.5cm} 2.55 \hspace{0.5cm} 12.30 & 2.9 \hspace{0.5cm} 15.43 \hspace{0.5cm} 72.89 \\
 
 ZmBART & \textbf{34.94} \hspace{0.5cm} \textbf{19.38} \hspace{0.5cm} \textbf{32.74}  & \textbf{4.4} \hspace{0.5cm} \textbf{26.51} \hspace{0.5cm} \textbf{74.19}  & \textbf{21.27} \hspace{0.5cm} \textbf{5.30} \hspace{0.5cm}\textbf{ 17.64} & \textbf{4.1} \hspace{0.5cm} \textbf{21.05} \hspace{0.5cm} \textbf{73.39} \\
 \hline 
 \multicolumn{5}{l}{\textit{Cross-lingual few-shot generation results (with 1000 supervised data points)}} \\\hline
 ZmBART & 52.37 \hspace{0.5cm} 35.52 \hspace{0.5cm} 50.50  & 7.6 \hspace{0.5cm} 34.11 \hspace{0.5cm} 78.29  & 36.29 \hspace{0.5cm} 14.21 \hspace{0.5cm} 27.22 & 6.5 \hspace{0.5cm} 26.58 \hspace{0.5cm} 78.27 \\
 \hline \hline
\end{tabular}}
\caption{\label{tab:hindi_auto_results} Zero and few-shot cross-lingual generation  results for Hindi Language}
\end{table*}

\begin{table*}
\centering
\scalebox{0.8}{
\begin{tabular}{l|c|c|c}
\hline\hline
\textbf{Model} & \textbf{News Headline Generation} & \textbf{Question Generation}& \textbf{Abstractive TS} \\
\hline
 \textbf{Metrics} & \textbf{R-1} \hspace{0.5cm} \textbf{R-2} \hspace{0.5cm}\textbf{ R-L} & \textbf{BL} \hspace{0.5cm} \textbf{R-L} \hspace{0.5cm} \textbf{BS} & \textbf{ R-1} \hspace{0.5cm} \textbf{R-2} \hspace{0.5cm} \textbf{R-L}  \\
\hline\hline
\multicolumn{4}{l}{\textit{Cross-lingual zero-shot generation results}} \\\hline
 MT Pipeline(mBART) & 13.82 \hspace{0.5cm} 0.38 \hspace{0.5cm} 7.92  & 8.9 \hspace{0.5cm} 26.92 \hspace{0.5cm} 71.93  & 17.90 \hspace{0.5cm} 3.98 \hspace{0.5cm} 18.46 \\
 
 mBART+\textsc{MADMo} & 33.75 \hspace{0.5cm} 8.12 \hspace{0.5cm} 17.78  & 16.6 \hspace{0.5cm} 34.80 \hspace{0.5cm} 74.01  & 28.74 \hspace{0.5cm} 9.01 \hspace{0.5cm} 23.63 \\
 
 mBART+\textsc{MADPd} & 31.58 \hspace{0.5cm} 6.98 \hspace{0.5cm} 18.95  & 18.2 \hspace{0.5cm} 36.22 \hspace{0.5cm} 74.99  & 19.17 \hspace{0.5cm} 4.89 \hspace{0.5cm} 18.22\\
 
 ZmBART & \textbf{35.25} \hspace{0.5cm} \textbf{9.24} \hspace{0.5cm} \textbf{19.92}  & \textbf{18.8} \hspace{0.5cm} \textbf{38.74} \hspace{0.5cm} \textbf{75.91}  & \textbf{36.60} \hspace{0.5cm} \textbf{15.26} \hspace{0.5cm}\textbf{29.85} \\
 \hline 
 \multicolumn{4}{l}{\textit{Cross-lingual few-shot generation results (with 1000 supervised data points)}} \\\hline
 ZmBART & 47.06 \hspace{0.5cm} 22.36 \hspace{0.5cm} 31.55  & 30.4 \hspace{0.5cm} 53.98 \hspace{0.5cm} 82.66  & 41.65 \hspace{0.5cm} 20.33 \hspace{0.5cm} 33.49 \\
 \hline \hline
\end{tabular}}
\caption{\label{tab:japanese_auto_results} Zero and few-shot cross-lingual generation results for Japanese Language}
\end{table*}

\begin{table*}
\centering
\scalebox{0.8}{
\begin{tabular}{l|c|c|c|c}
\hline\hline
\textbf{Model} & \textbf{News Headline Generation} & \textbf{Question Generation}& \textbf{Abstractive TS} & \textbf{Distractor Generation}\\
\hline
 \textbf{Metrics} & \textbf{Flu} \hspace{0.5cm} \textbf{Rel} \hspace{0.5cm}\textbf{Corr} & \textbf{Flu} \hspace{0.5cm} \textbf{Rel} \hspace{0.5cm} \textbf{Corr} & \textbf{Flu} \hspace{0.5cm} \textbf{Rel} \hspace{0.5cm} \textbf{Corr} & \textbf{Flu} \hspace{0.5cm} \textbf{Rel} \hspace{0.5cm}\textbf{Dist} \\
\hline\hline
\multicolumn{5}{l}{\textit{Annotator set-01}} \\\hline
 mBART+\textsc{MADMo} & 3.86 \hspace{0.5cm} 4.34 \hspace{0.5cm} 3.94  & 2.66 \hspace{0.5cm} 3.38 \hspace{0.5cm} 3.52  & 3.56 \hspace{0.5cm} 3.58 \hspace{0.5cm} 3.22 & 3.61 \hspace{0.5cm} 4.08 \hspace{0.5cm} 2.89 \\
 mBART+\textsc{MADPd} & 2.54 \hspace{0.5cm} 2.96 \hspace{0.5cm} 2.28  & 3.1 \hspace{0.5cm} 3.4 \hspace{0.5cm} 3.78  & 2.26 \hspace{0.5cm} 2.62 \hspace{0.5cm} 1.92 & 2.42 \hspace{0.5cm} 3.72 \hspace{0.5cm} 3.08 \\
 ZmBART & \textbf{4.14} \hspace{0.5cm} \textbf{4.22} \hspace{0.5cm} \textbf{4.04}  & \textbf{3.24} \hspace{0.5cm} \textbf{3.44} \hspace{0.5cm} \textbf{3.9}  & \textbf{4.02} \hspace{0.5cm} \textbf{4.12} \hspace{0.5cm}\textbf{3.54} & \textbf{4.12} \hspace{0.5cm} \textbf{4.19} \hspace{0.5cm} \textbf{3.83} \\
 \hline 
 \multicolumn{5}{l}{\textit{Annotator set-02}} \\\hline
 mBART+\textsc{MADMo} & 3.84 \hspace{0.5cm} 4.18 \hspace{0.5cm} 3.8  & 3.83 \hspace{0.5cm} 4.63 \hspace{0.5cm} 3.96  & 3.38 \hspace{0.5cm} 3.96 \hspace{0.5cm} 3.4 & 3.38 \hspace{0.5cm} 3.00 \hspace{0.5cm} 2.24 \\
 mBART+\textsc{MADPd} & 2.96 \hspace{0.5cm} 3.02 \hspace{0.5cm} 2.7  & \textbf{3.98} \hspace{0.5cm} 4.70 \hspace{0.5cm} 3.98  & 2.96 \hspace{0.5cm} 3.16 \hspace{0.5cm} 2.84 & 2.97 \hspace{0.5cm} 3.11 \hspace{0.5cm} \textbf{2.46} \\
 ZmBART & \textbf{4.12} \hspace{0.5cm} \textbf{4.38} \hspace{0.5cm} \textbf{4.16}  & 3.95 \hspace{0.5cm} \textbf{4.80} \hspace{0.5cm} \textbf{4.27}  & \textbf{4.24} \hspace{0.5cm} \textbf{4.52} \hspace{0.5cm}\textbf{ 4.38} & \textbf{3.56} \hspace{0.5cm} \textbf{3.18} \hspace{0.5cm} 2.36 \\
 \hline
 \multicolumn{5}{l}{\textit{Annotator set-03}} \\\hline
 mBART+\textsc{MADMo} & 3.56 \hspace{0.5cm} 3.74 \hspace{0.5cm} \textbf{3.78}  & 2.68 \hspace{0.5cm} 3.76 \hspace{0.5cm} 3.32  & 2.9 \hspace{0.5cm} 3.34 \hspace{0.5cm} 2.9 & 3.96 \hspace{0.5cm} 3.74 \hspace{0.5cm} 3.12 \\
 mBART+\textsc{MADPd} & 3.1 \hspace{0.5cm} 3.42 \hspace{0.5cm} 2.91  & 2.80 \hspace{0.5cm} 3.88 \hspace{0.5cm} 3.56  & 2.64 \hspace{0.5cm} 2.34 \hspace{0.5cm} 2.46 & 4.13 \hspace{0.5cm} 3.74 \hspace{0.5cm} 2.94 \\
 ZmBART & \textbf{3.70} \hspace{0.5cm} \textbf{3.84} \hspace{0.5cm} 3.76  & \textbf{2.86} \hspace{0.5cm} \textbf{4.04} \hspace{0.5cm} \textbf{3.76}  & \textbf{4.06} \hspace{0.5cm} \textbf{3.56} \hspace{0.5cm}\textbf{ 3.56} & \textbf{4.44} \hspace{0.5cm} \textbf{4.12} \hspace{0.5cm} \textbf{3.12} \\
 \hline \hline
\end{tabular}}
\caption{\label{tab:hindi_man_results} Manual evaluation results of Zero-shot generated outputs for Hindi language}
\end{table*}


\begin{table*}
\centering
\scalebox{0.8}{
\begin{tabular}{l|c|c|c}
\hline\hline
\textbf{Model} & \textbf{News Headline Generation} & \textbf{Question Generation}& \textbf{Abstractive TS} \\
\hline
 \textbf{Metrics} & \textbf{Flu} \hspace{0.5cm} \textbf{Rel} \hspace{0.5cm}\textbf{Corr} & \textbf{Flu} \hspace{0.5cm} \textbf{Rel} \hspace{0.5cm} \textbf{Corr} & \textbf{Flu} \hspace{0.5cm} \textbf{Rel} \hspace{0.5cm} \textbf{Corr}  \\
\hline\hline
\multicolumn{4}{l}{\textit{Annotator set-01}} \\\hline
 mBART+\textsc{MADMo} & 2.66 \hspace{0.5cm} 2.98 \hspace{0.5cm} 2.50 & 1.98 \hspace{0.5cm} \textbf{3.70} \hspace{0.5cm} \textbf{3.18} & 3.04 \hspace{0.5cm} 3.55 \hspace{0.5cm} 3.44  \\
 mBART+\textsc{MADPd} & 2.26 \hspace{0.5cm} 2.70 \hspace{0.5cm} 2.04  & 2.00 \hspace{0.5cm}  3.38   \hspace{0.5cm} 2.82  & 1.44 \hspace{0.5cm} 2.22 \hspace{0.5cm} 2.20 \\
 ZmBART & \textbf{3.60} \hspace{0.5cm} \textbf{4.02} \hspace{0.5cm} \textbf{3.50}  & \textbf{2.12} \hspace{0.5cm} 3.30 \hspace{0.5cm} 2.94 & \textbf{4.24} \hspace{0.5cm} \textbf{3.90} \hspace{0.5cm} \textbf{3.90}   \\
 \hline 
 \multicolumn{4}{l}{\textit{Annotator set-02}} \\\hline
 mBART+\textsc{MADMo} & 2.1 \hspace{0.5cm} 2.58 \hspace{0.5cm} 1.98 & 1.24 \hspace{0.5cm} 1.70 \hspace{0.5cm}  1.33 & 2.56 \hspace{0.5cm} 3.40 \hspace{0.5cm} 2.62 \\
  mBART+\textsc{MADPd} & 1.58 \hspace{0.5cm} 1.78 \hspace{0.5cm} 1.46  & \textbf{1.46} \hspace{0.5cm} 1.72 \hspace{0.5cm}  1.78 & 1.00 \hspace{0.5cm} 1.00 \hspace{0.5cm} 1.00 \\
 ZmBART & \textbf{3.78} \hspace{0.5cm} \textbf{4.16} \hspace{0.5cm} \textbf{3.86}  & 1.26 \hspace{0.5cm} \textbf{1.76} \hspace{0.5cm} \textbf{1.88}  & \textbf{4.04} \hspace{0.5cm} \textbf{4.26} \hspace{0.5cm}\textbf{3.84} \\
 \hline

 \multicolumn{4}{l}{\textit{Annotator set-03}} \\\hline
mBART+\textsc{MADMo} & 2.24 \hspace{0.5cm} 2.72 \hspace{0.5cm} 2.24  & \textbf{2.34} \hspace{0.5cm}  2.46 \hspace{0.5cm} 2.39  & 2.82 \hspace{0.5cm} 3.18 \hspace{0.5cm} 3.52 \\

  mBART+\textsc{MADPd} & 1.9 \hspace{0.5cm} 2.14 \hspace{0.5cm} 1.82 & 2.10 \hspace{0.5cm} 2.66 \hspace{0.5cm} 2.28  & 1.16 \hspace{0.5cm}  1.84 \hspace{0.5cm} 1.44 \\

 ZmBART & \textbf{2.88} \hspace{0.5cm} \textbf{3.22} \hspace{0.5cm} \textbf{2.92}  & 2.10 \hspace{0.5cm} \textbf{2.70} \hspace{0.5cm} \textbf{2.46 } & \textbf{3.32} \hspace{0.5cm} \textbf{3.52} \hspace{0.5cm}\textbf{3.04} \\
 
 \hline\hline
\end{tabular}}
\caption{\label{tab:japanese_man_results} Manual evaluation results of Zero-shot generated outputs for Japanese language}
\end{table*}
\vspace{-3mm}
\subsection{News Headline Generation (NHG)}
\label{subsec:nhg}
In this task, \textit{given a news article, we generate grammatically coherent, semantically correct and abstractive headline}. We use 500k/30k/30k (train/validation/test) English NHG data splits from \textit{Gigaword} headline generation corpus\footnote{\url{https://github.com/harvardnlp/sent-summary}}. For Hindi and Japanese we use 1k/1k/5k spilt from Kaggle\footnote{\url{https://www.kaggle.com/disisbig/hindi-text-short-summarization-corpus}} (we manually filtered high-quality news and headlines) and \cite{iwamakano2019multiple} respectively. 

In a zero-shot setting we fine-tune ZmBART model on supervised data and directly evaluate results on Hindi and Japanese test datasets. Automated evaluation results are included in Tables \ref{tab:hindi_auto_results} and \ref{tab:japanese_auto_results}. We observe that, quality of generated headlines in Hindi is better compared to Japanese. The possible reasoning can be the input size. 
ZmBART outperforms the baseline with an absolute difference of 5.22 ROUGE-L score. mBART+\textsc{MADMo} is best among others which shows that masking and denoising with monolingual data indeed enrich the multi-lingual latent space for selected three languages. mBART+\textsc{MADMo} generates code mixed (Hindi-English or Hindi-Japanese) output which degrades the model performance. Few-shot training fills the mistakes of zero-shot models and generates better quality output. Manual evaluation scores (Tables \ref{tab:hindi_man_results} and \ref{tab:japanese_man_results}) and automated scores correlate well validating ZmBART's performance on NHG task.  

\vspace{-3mm}
\subsection{Question Generation (QG)}
\label{subsec:qg}
In the Question Generation (QG) task,  \textit{given an input passage and an answer, the aim is to generate semantically and syntactically correct questions that can produce the answer.} We use SQuAD 1.1 \cite{rajpurkaretal2016squad} English data for supervised training. SQuAD is popular question answering dataset consisting of 100k+ $<$passage, question, answer$>$ tuples. Following \cite{zhaoetal2018paragraph}, we combine the train and validation sets of SQuAD and then spilt it as 80k/8k/10k training/validation/test tuples. For Hindi we use 1k/5.5k (train/test) from MLQA \cite{lewisetal2020mlqa} and TyDiQA-GoldP \cite{Clark2020tydiqa} datasets. We use 1k/1k/5k for Japanese data from \cite{takahashietal2019machine}. Hindi and Japanese data are available in SQuAD data format which maintains consistency in terms of passage size, question and number of answers. For given passage and question we randomly selected one answer to form the dataset. We combine answer and passage as single input sequence separated by special token $<$s$>$.

 Even without any parallel data, ZmBART outperformed all the baselines consistently across all automated evaluation metrics for zero-shot setting. Regarding manual evaluations, we see that Hindi questions received good score from the annotators, whereas the questions generated for the Japanese language inputs were considered as poor. Upon closer inspection of the generated text we find that several generated questions start with English wh-words. This mixing of English 'code' in the output happened somewhat seamlessly for the Hindi data as tokens in both languages are written in left-to-right manner. Moreover, Hindi-English code-mixed data is now getting very common and the annotators mostly accepted the mixing of the wh-words with the Hindi texts. Such mixing is not very common with Japanese text. As a result, the annotators assigned lower scores to such texts. 

We then tried to understand the reason for getting the wh-words at the beginning of the output. English interrogative sentences often introduce wh-words at the beginning even though they are not present in the original data. 
The model gets exposed to such special characteristics of the English interrogative sentences during the fine tuning. The output from other languages get impacted due to this in zero-shot settings. However, the semantics of the text is captured well for the model as demonstrated by the high BERTScore, indicating good cross-lingual transfer of semantic knowledge.

\vspace{-.1in}
\subsection{Abstractive Text Summarization (ATS)}
\label{subsec:ats}
In Abstractive Text Summarization (ATS), we aim to \textit{generate grammatically coherent, semantically correct and abstractive summary given an input document}. We use recently released WikiLingua \cite{ladhaketal2020wikilingua} cross-lingual abstractive summarization dataset containing data in 18 languages. 
Prior splits are not available for this dataset. We use 131k/5k/5k (train/validation/test) splits for English, and 1k/1k/5k splits for Hindi and Japanese. 

By skimming through data in Hindi we observe that many input documents consist of technical instructions on usage of softwares/tools. Summarizing these instructions are challenging. Zero-shot ZmBART performed  better as compared to baselines as shown in human evaluation (Tables \ref{tab:hindi_man_results} and \ref{tab:japanese_man_results} for Hindi and Japanese respectively). The human evaluation results correlate with automated evaluation as shown in Tables \ref{tab:hindi_auto_results} and \ref{tab:japanese_auto_results}. 
\citet{ladhaketal2020wikilingua} reported cross-lingual ATS score with same data for four different languages. The R-L score for four languages are 34.06, 37.09, 31.67 and 32.33. We obtain R-L scores of 27.22 and 33.49 for Hindi and Japanese respectively, which shows that the few-shot performance of ZmBART is acceptable.

\vspace{-.1in}
\subsection{Distractor Generation (DG)}
\label{subsec:dg}
The final task to judge ZmBART's performance is Distractor Generation (DG). It is the task of generating incorrect options (also known as distractors) from reading comprehension MCQ. The generated distractors should be in the context with the question but shouldn't be semantically equivalent to the answer. Formally, \textit{for given passage, question and answer triplet, generate a long, coherent, and grammatically correct wrong option}. Considering the fact that for a given triplet there can be many incorrect options that are completely different from each other, the problem is even more challenging. We use English DG dataset from \cite{maurya2020learning} which consists of approx 135k/17k/17k (train/validation/test) split. We were unable to find a suitable dataset in Japanese language. For Hindi language we created a dataset called \textbf{\textit{HiDG}}\footnote{Implementation, dataset, pre-trained checkpoints and ZmBART generated text are available at \url{ https://github.com/kaushal0494/ZmBART}} of 1k/1k/5k split. Similar to QG, to create input for ZmBART we concatenate the answer, question and passage in the same order and separate them with special token $<$s$>$. 

To generate HiDG, we first extracted $<$passage, question, answer$>$ triplets from English SQuAD 1.1 with atleast 150 tokens in the triplet. We generate distractors for these examples using model proposed by \citet{maurya2020learning}. The distractors were translated to Hindi using Google Translator service. The translated distractors were manually verified or corrected (if necessary) by human annotators.

The evaluation of the task is challenging because: 1) there can be more then one correct distractors. Automated evaluation metrics may not able to capture this aspect as only one ground truth distractor is available and 2) it may possible that the generated distractor is semantically similar to answer with high lexical  overlap with reference distractor in those situation lexical match based metrics are not suitable. To evaluate the DG task we mainly rely on BERTScore and manual evaluation. Towards this effort we consider higher number of DG samples for manual evaluation.  Results from Tables \ref{tab:hindi_auto_results} and \ref{tab:hindi_man_results} indicate the superiority of ZmBART over the baseline models for this task.

To summarize, we have performed experiments for 14 different task-setup combinations involving low resource languages. With four tasks in Hindi and three tasks in Japanese, and each task in zero shot and few shot setup, we provide detailed comparative evaluation for the tasks. The tasks are of different natures, and each task offers its own unique challenge. We critically analyze the performances to show the robustness and the range of applicability for the proposed ZmBART framework. We use fairseq library \cite{ott2019fairseq} for all the implementation and experiments. The implementation details are included in supplementary.
\vspace{-7mm}
\section{Results Analysis and Ablation Study}
\label{sec:abandanlys}
In this section, we provide further analysis of the experimental results. We also perform ablation studies to understand the impacts of the different modeling decisions made in designing the framework.

\textbf{\textbullet Supervised Training Results: }
\label{subsec:supresult}
Table \ref{tab:sup_results} shows the comparative results of fine-tuned mBART with and without auxiliary task on task-specific supervised English data. We observe that there is no significant performance degradation of ZmBART over original mBART model with pure supervised training. Even, the auxiliary task helps in achieving slight improvement over the original mBART performance in most setups. This concludes that ZmBART can be adopted as replacement of original mBART model with additional functionalities.

\begin{table}[!htb]
\centering
\scalebox{0.7}{
\begin{tabular}{lcccccc}
\hline \hline  \textbf{Task} & \textbf{Setting} & \textbf{BL} & \textbf{R-1} & \textbf{R-2} & \textbf{R-L} & \textbf{BS} \\ \hline
NHG & W/ Aux-Task & 15.9 & 43.22 & 21.33 & 40.88 & 90.13 \\
 & W/O Aux-Task & 15.9 & 43.15 & 21.25 & 40.77 & 90.13 \\
\hline

QG & W/ Aux-Task &  20.6 & 53.20 & 26.53 & 51.37 & 92.18 \\
& W/O Aux-Task & 21.4 & 52.66 & 26.63 & 51.25 & 92.41  \\
\hline

ATS & W/ Aux-Task & 16.0  & 40.01 & 18.11 & 38.29  & 90.20 \\
& W/O Aux-Task & 15.8  & 39.52 & 18.00 & 37.91 & 90.10 \\
\hline

DG & W/ Aux-Task & 10.3  & 31.76 & 14.89 & 31.18 & 89.33 \\
& W/O Aux-Task & 10.0  & 31.87 & 14.59 & 31.30 & 89.42  \\
\hline \hline
\end{tabular}}
\caption{\label{tab:sup_results} Automated evaluation results of mBART on task-specific supervised English dataset (with and without Auxiliary Task)}
\vspace{-.2in}
\end{table} 

\textbf{\textbullet Effect of Auxiliary Task: }
\label{subsec:effect_aux}
Table \ref{tab:eff_auxi_task} includes the results with and without auxiliary task of ZmBART for ATS and QG tasks in zero-shot setting. It can be inferred that without the auxiliary task, lexical match based scores are poor because the decoder generates code-mixed outputs. We see that the BERTScore is still reasonable without auxiliary task owing to the multilingual mBART embedding. However, generation of the data in appropriate language is enabled only after inclusion of the auxiliary task. The auxiliary task contributes in two ways: it enables zero-shot generation and  improves the mBART multilingual latent space even more as indicated by the improved BERTScore.

\begin{table}[!htb]
\centering
\scalebox{0.6}{
\begin{tabular}{l|c|c}
\hline\hline
\textbf{Model} & \textbf{Abstractive TS}& \textbf{Question Generation} \\
\hline
 \textbf{Metrics}  & \textbf{R-1} \hspace{0.5cm} \textbf{R-2} \hspace{0.5cm} \textbf{R-3} & \textbf{BL} \hspace{0.5cm} \textbf{R-L} \hspace{0.5cm} \textbf{BS}  \\
\hline\hline
\multicolumn{3}{l}{\textit{Hindi Language}} \\\hline
 ZmBART w/o Aux  & 4.34 \hspace{0.5cm} 0.10 \hspace{0.5cm} 3.19  & 0.9 \hspace{0.5cm} 16.64 \hspace{0.5cm} 70.72 \\
 
 ZmBART with Aux  & 21.27 \hspace{0.5cm} 5.30 \hspace{0.5cm} 17.64 & 4.4 \hspace{0.5cm} 26.51 \hspace{0.5cm} 74.19 \\
 \hline 
 \multicolumn{3}{l}{\textit{Japanese Language}} \\\hline
 ZmBART w/o Aux  & 6.80 \hspace{0.5cm} 0.11 \hspace{0.5cm} 5.30  & 6.7 \hspace{0.5cm} 33.07 \hspace{0.5cm} 70.35 \\
 
 ZmBART with Aux  & 36.60 \hspace{0.5cm} 15.26 \hspace{0.5cm} 29.89 & 18.8 \hspace{0.5cm} 38.74 \hspace{0.5cm} 75.91 \\
 \hline\hline
\end{tabular}}
\caption{\label{tab:eff_auxi_task} Zero-shot results of ZmBART with and without auxiliary task for Hindi and Japanese}
\vspace{-.3in}
\end{table}

With these results we now want to understand \emph{whether the auxiliary task is able to generalize across multiple tasks, or favors specific tasks.} Among the tasks considered in this work, we see that generation of meaningful summaries/headlines require understanding/abstracting of input text which is unlikely to be obtained by repeating sentences from input passages, as done in the auxiliary task. ZmBART achieves good zero-shot/few-shot/supervised results (Tables 1-5) on ATS and NHG over strong baselines. The generated headlines and summaries were found to be mostly abstractive, they don't contain large continuous sequences from input text. As described in Sections \ref{subsec:qg} and \ref{subsec:dg}, Question Generation and Distractor Generation are more challenging tasks and have objectives vastly different from the auxiliary task’s objective. Even for these tasks, decent evaluation scores (Tables 1-5) and improvements over the baselines across the languages considered indicate that the solutions are not spurious. Incorporation of auxiliary task improves the performance of diverse downstream tasks on real benchmark datasets, and does not favor any specific task or dataset.

\textbf{\textbullet~ Approaches to avoid Catastrophic Forgetting:}
We use two approaches to address the catastrophic forgetting problem, (a) Freezing model components and (b) optimized regularization  \cite{van2019three}. Tables \ref{tab:cats_effect_qg} and \ref{tab:cats_effect_nhg} show the automated evaluation results with different approaches used to deal with the catastrophic forgetting problem.  It can be noted that the proposed modelling setup (i.e., ZmBART) gives best results. 

\begin{table*}[!htb]
\centering
\scalebox{0.8}{
\begin{tabular}{llccc}
\hline \hline  \textbf{Setup} & \textbf{Setting-Details} & \textbf{BL(hi/ja)} & \textbf{R-L(hi/ja)} & \textbf{BS(hi/ja)} \\ \hline Model Components & Freeze word embedding (WE) & 2.5/13.6  & 21.55/31.99 & 72.02/73.18 \\
 & Freeze WE + subset of Encoder \& Decoder layers & 2.9/15.3  & 22.62/36.60 & 72.24/72.98 \\
 & Freeze WE + Encoder layers & 2.2/13.8  & 19.69/36.91 & 69.73/72.97 \\
 & Freeze WE + Decoder layers (ZmBART) & \textbf{4.4/18.8}  & \textbf{26.51/38.74} & \textbf{74.19/75.91} \\ 
\hline

Regularized Optimization & Elastic Weight Consolidation (EWC) &  2.1/11.6 & 18.21/29.47 & 68.36/72.91 \\

\hline \hline
\end{tabular}
}
\caption{\label{tab:cats_effect_qg} Evaluation scores for different modeling approaches to avoid  catastrophic-forgetting for QG Task}
\vspace{-.1in}
\end{table*}

\begin{table*}[!htb]
\centering
\scalebox{0.8}{
\begin{tabular}{llccc}
\hline \hline  \textbf{Setup} & \textbf{Setting-Details} & \textbf{R-1(hi/ja)} & \textbf{R-2(hi/ja)} & \textbf{R-L(hi/ja)} \\ \hline Model Components & Freeze word embedding (WE) & 13.02/26.07  & 05.67/03.96 & 12.45/17.62 \\
 & Freeze WE + subset of Encoder \& Decoder layers & 14.27/25.72  & 06.70/03.21 & 13.76/18.28 \\
 & Freeze WE + Encoder layers & 09.81/22.67  & 04.10/02.38 & 09.66/13.68 \\
 & Freeze WE + Decoder layers (ZmBART) & \textbf{34.94/35.25}  & \textbf{19.38/09.24} & \textbf{32.74/19.92} \\ 
\hline

Regularized Optimization & Elastic Weight Consolidation (EWC) &  12.01/22.16 & 05.43/03.11 & 11.22/16.31 \\

\hline \hline
\end{tabular}
}
\caption{\label{tab:cats_effect_nhg} \small Evaluation scores for different modeling approaches to avoid catastrophic-forgetting for NHG Task}
\vspace{-.1in}
\end{table*} 

\textbf{\textbullet~ Effect of Architecture on Few-shot Training: }
\label{subsec:few_sup}
In this set-up we experiment with few-shot training with mBART (directly fine-tuned on task-specific supervised English data) and ZmBART (trained with auxiliary task and fine-tuned with English data). 
The results are presented in Table \ref{tab:eff_few_arc}. We find that ZmBART does better than mBART in corresponding setups. Moreover, although freezing the decoder layer and word embeddings helps in zero-shot setting, it is natural and useful to unfreeze them during few shot training. 

\begin{table}[!htb]
\centering
\scalebox{0.7}{
\begin{tabular}{l|c|c}
\hline\hline
\textbf{Model} & \textbf{NHG} & \textbf{QG} \\
\hline
 \textbf{Metrics}  & \textbf{R-1} \hspace{0.5cm} \textbf{R-2} \hspace{0.5cm} \textbf{R-3} & \textbf{BL} \hspace{0.5cm} \textbf{R-L} \hspace{0.5cm} \textbf{BS} \\
\hline\hline
 mBART+WE & 50.61 \hspace{0.5cm} 34.32 \hspace{0.5cm} 49.01 & 6.1 \hspace{0.5cm} 31.20 \hspace{0.5cm} 77.01 \\
 mBART  & 51.49 \hspace{0.5cm} 35.04 \hspace{0.5cm} 49.64  & 7.1 \hspace{0.5cm} 32.96 \hspace{0.5cm} 77.61 \\
 ZmBART+WE  & 51.81 \hspace{0.5cm} 35.04 \hspace{0.5cm} 50.07  & 6.9 \hspace{0.5cm} 32.82 \hspace{0.5cm} 77.40 \\
 ZmBART  & \textbf{52.37} \hspace{0.5cm} \textbf{35.52} \hspace{0.5cm} \textbf{50.50}  & \textbf{7.9} \hspace{0.5cm} \textbf{34.49} \hspace{0.5cm} \textbf{78.39}\\
 \hline\hline
\end{tabular}}
\caption{\label{tab:eff_few_arc} \small Hindi language few-shot results for different architectural setups. \textit{WE} indicates that word embeddings and decoder layer parameters are frozen}
\vspace{-.2in}
\end{table}

\begin{figure}[!htb]
\begin{subfigure}
    \centering
    \includegraphics[width=8cm]{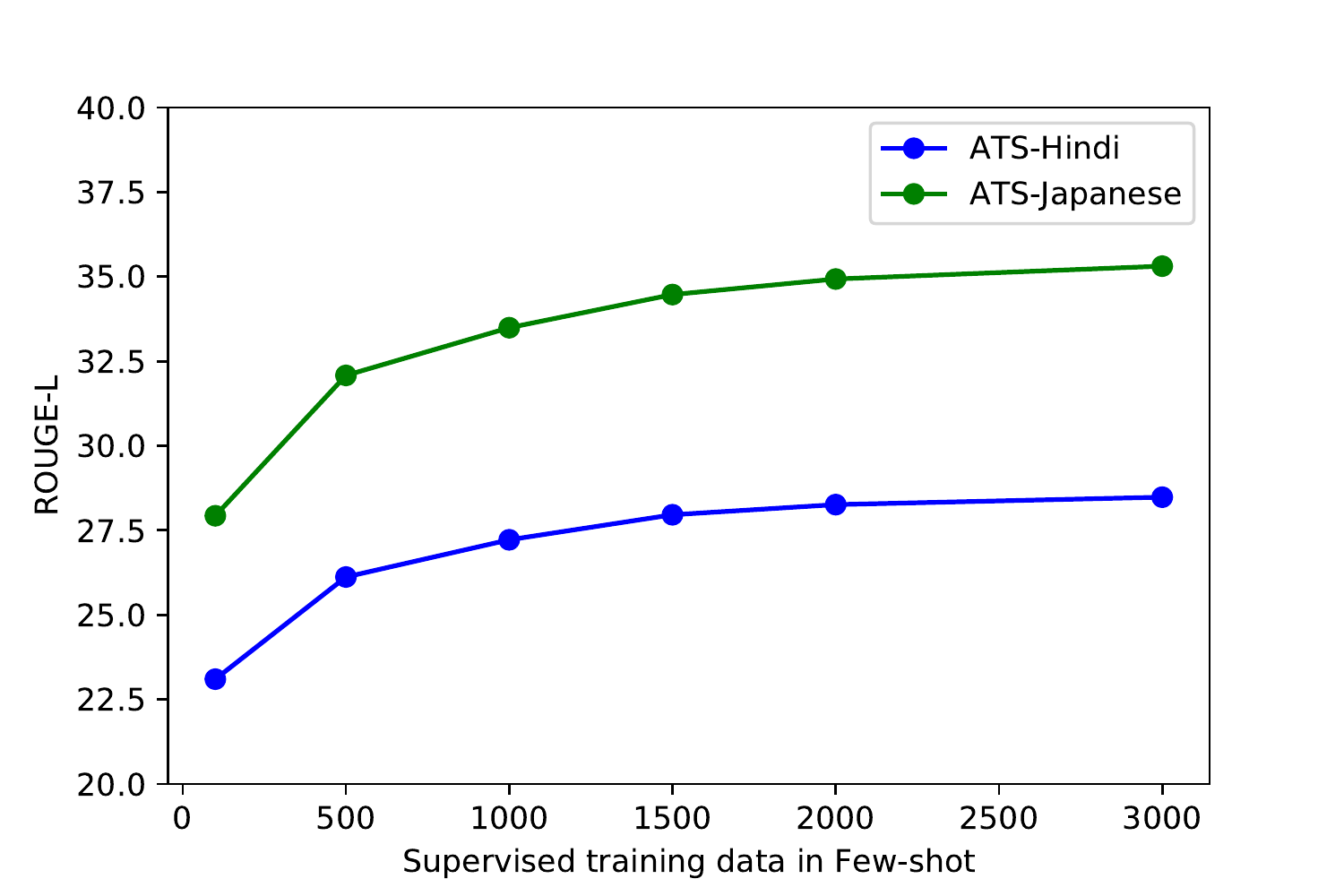}
    \vspace{-.23in}
    \caption{\small ZmBART model few-shot performance with supervised Hindi/Japanese data for ATS task}
    \label{fig:effect_Sup1}
\end{subfigure}
\vspace{-.1in}
\begin{subfigure}
    \centering
    \includegraphics[width=8cm]{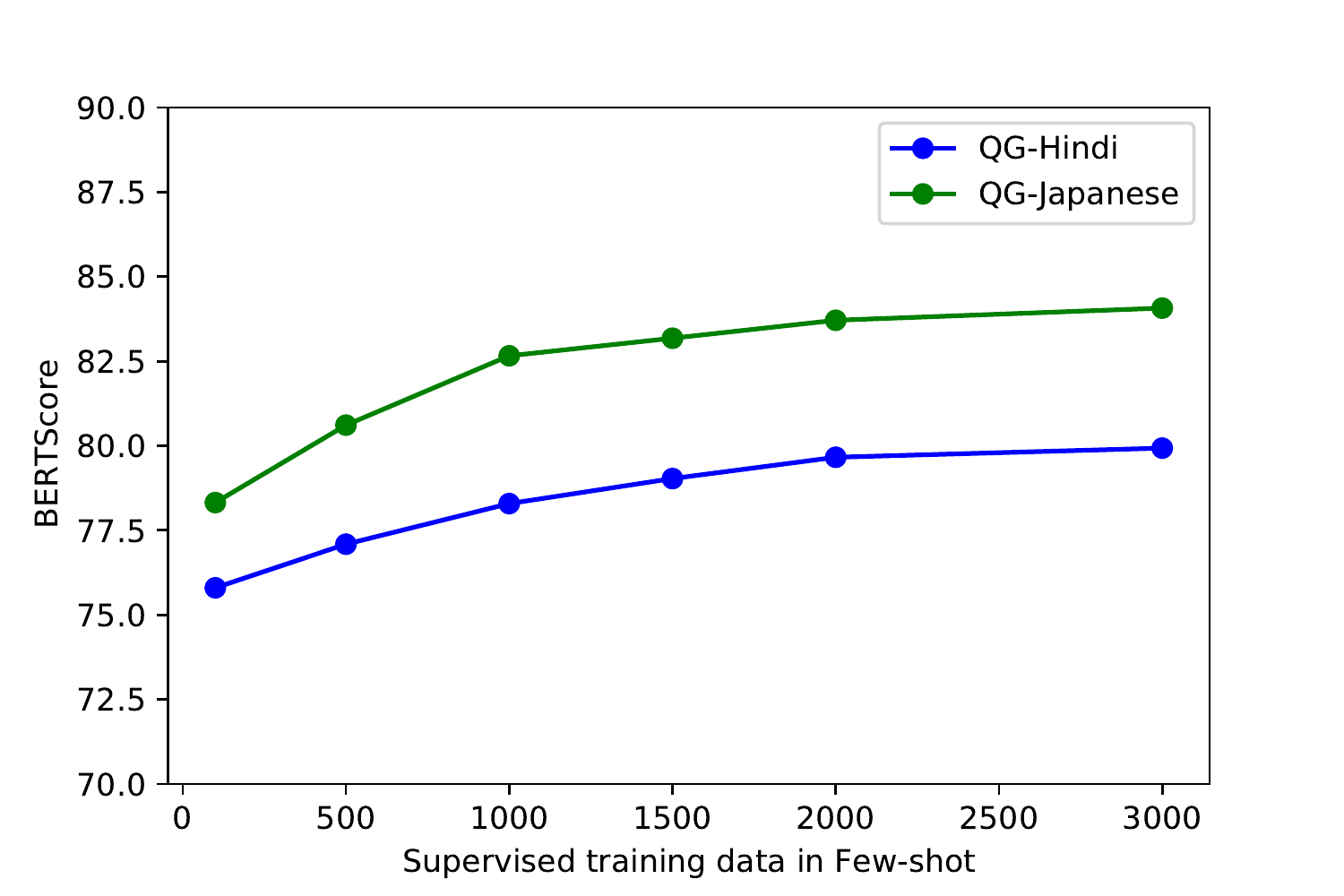}
    \vspace{-.23in}
    \caption{\small ZmBART model few-shot performance with supervised Hindi/Japanese data for QG task}
    \label{fig:effect_Sup2}
\end{subfigure}
\vspace{-.1in}
\end{figure}

\textbf{\textbullet~ Few-shot performance with Supervised data: } 
\label{subsec:few_sup}
Figures \ref{fig:effect_Sup1} and \ref{fig:effect_Sup2} show the trends of few-shot training of ZmBART with respect to supervised Hindi and Japanese training data for ATS and QG tasks respectively. We observe that with a small number of supervised examples (e.g. 100) the model achieves decent few-shot performance. We found the trends for different tasks to be similar. The improvement in model performance tends to be minimal after 1000 examples.

\vspace{-.1in}


\section{Conclusion}
\label{sec:con}
\vspace{-.1in}
In this paper, we propose a novel unsupervised  framework (ZmBART) for cross-lingual transfer and generation. The framework transfers supervision from HRL to LRLs which enables zero-shot language generation. The framework does not use any direct or pseudo-parallel data. ZmBART is directly applied to multiple generation tasks and languages. The model includes a carefully designed auxiliary task that further improved the multilingual embedding space, and helped to initialize encoder-decoder weights to enable zero shot language generation. We performed experiments in three languages and 18 task-setup combinations: four supervised tasks in English, four tasks in Hindi (each with zero-shot and few-shot), and three tasks in Japanese (each with zero-shot and few-shot). Except zero-shot question generation tasks, for all other tasks involving LRLs, the proposed model generated good quality results as validated by automated and manual evaluation measures. In future we want to extend this work by adding multiple other languages and tasks, and also explore other choices of auxiliary tasks for better model transfer.


\section*{Acknowledgments}
\vspace{-.1in}
We thank the support from Nvidia AI Technology Center (NVAITC) towards the requirements of computing power and compute infrastructure. We thank the human annotators for human evaluation and the anonymous reviewers for their constructive feedback.

\bibliographystyle{acl_natbib}
\bibliography{anthology,acl2021}
\clearpage
\section{Supplementary Materials}
\label{sec:supplementary}

\subsection{Implementation Details:}
We use a standard sequence-to-sequence Transformer architecture with 12 layers (each 16 heads) for encoder and decoder. The model has a dimension of 1024 (approx 680M parameters). Additional layer-normalization was used with both the encoder and decoder. We found FP16 precision stabilized the training. We trained all the models on 4 Nvidia V100 GPUs (32GB). Similar to mBART we use the Adam optimizer ($\epsilon= 1e\mhyph6$, $\beta_2 = 0.98$) and linear learning rate decay scheduling. The training started with a dropout value 0.3 and was later reduced to 0.2 after 20k steps and 0 after 40k steps. The loss function was cross-entropy label smoothing loss. 2500 warm-up steps and  $3e\mhyph5$ learning rate were used. The model selection was done based on validation data likelihood. We use beam-search with beam size 5 in the decoding for all the tasks. We loaded mBARTCC25 pre-trained checkpoint weights and further pre-train/fine-tune model on task-specific data with teacher forcing method. 

The above set of parameters are used for all the target tasks as well as the auxiliary task. We process different batch sizes of input for different tasks.  We use 2048, 3000, 4096, 2048, and 5000 tokens per GPU for ATS, DG, QG, auxiliary, and NHG tasks, respectively.  We use shared Byte Pair Encoding (BPE) vocabulary from sentence-piece tokenizer of size 250k. We use 34k/1k/1k (train/validation/test) data-points for auxiliary language (approx 11333 from each  languages). We train the mBART model with the auxiliary task around 10k steps. Training time for the auxiliary task is around 2-3 hours. The fine-tuning times for TS, QG, NHG, and DG were around 4-5, 1-2, 1-2, and 2-3 hours. We observe a longer fine-tuning time for ATS because of long passages. We selected the best model based on loss and perplexity on the validation datasets.  We checked with early-stopping and other checkpoints, which resulted in poor performance. 

\subsection{Evaluation  Metric and Tokenizer Details:}
For Automated evaluation, we use sacreBLEU implementation, ROUGE-L, and BERTScore. For ATS and NHG tasks, ROUGE-1, ROUGE-2, and ROUGE-L are used.  We explicitly use community-adopted language specific-tokenizers.  
Links for language-specific tokenizers are given below:
\begin{itemize}
    \item \textbf{English:} Default  sacreBLEU tokenizer i.e, \url{https://github.com/mjpost/sacrebleu}
    \item \textbf{Hindi:} \url{https://anoopkunchukuttan.github.io/indic_nlp_library/}
    \item \textbf{Japanese:} \url{http://www.phontron.com/kytea/}
\end{itemize}

Links of publicly available implementations of automated evaluation metrics which we use directly in this work:
\begin{itemize}
    \item \textbf{BLEU:} \url{https://github.com/mjpost/sacrebleu}
    \item \textbf{ROUGE:} \url{https://github.com/pltrdy/files2rouge}
    \item \textbf{BERTScore:} \url{https://github.com/Tiiiger/bert_score}
\end{itemize}

\subsection{Few Zero-shot Generated outputs from ZmBART:}
In the next few figures, we present sample outputs generated by the model in zero-shot setups, for Hindi and Japanese languages.

\label{subsec:few_sup}
\begin{figure*}
    \centering
    \includegraphics[width=11.7cm]{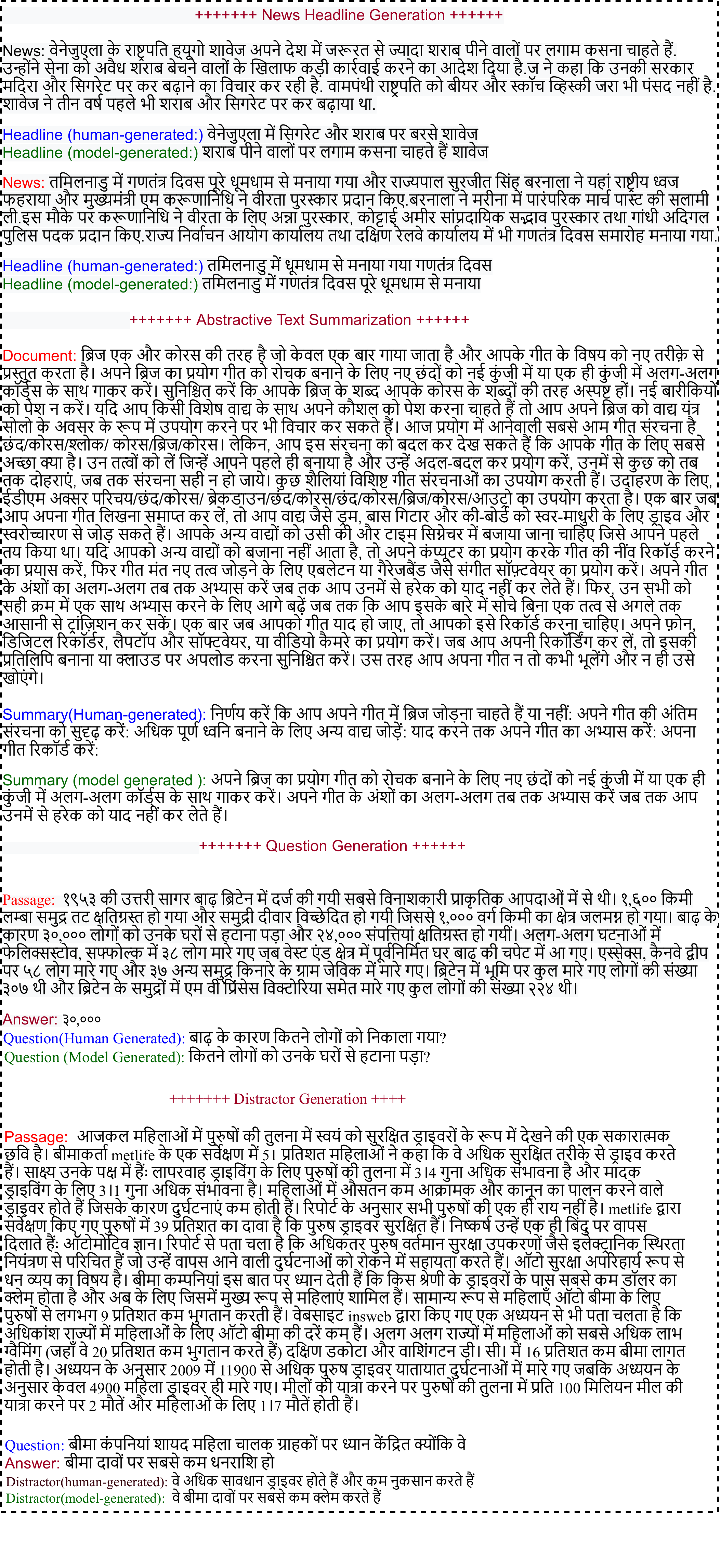}
    \label{fig:hindi_samp}
    \caption{Sample outputs for zero-short NHG, ATS, QG and DG in Hindi language}
\end{figure*}

\begin{figure*}
    \centering
    \includegraphics[width=14cm]{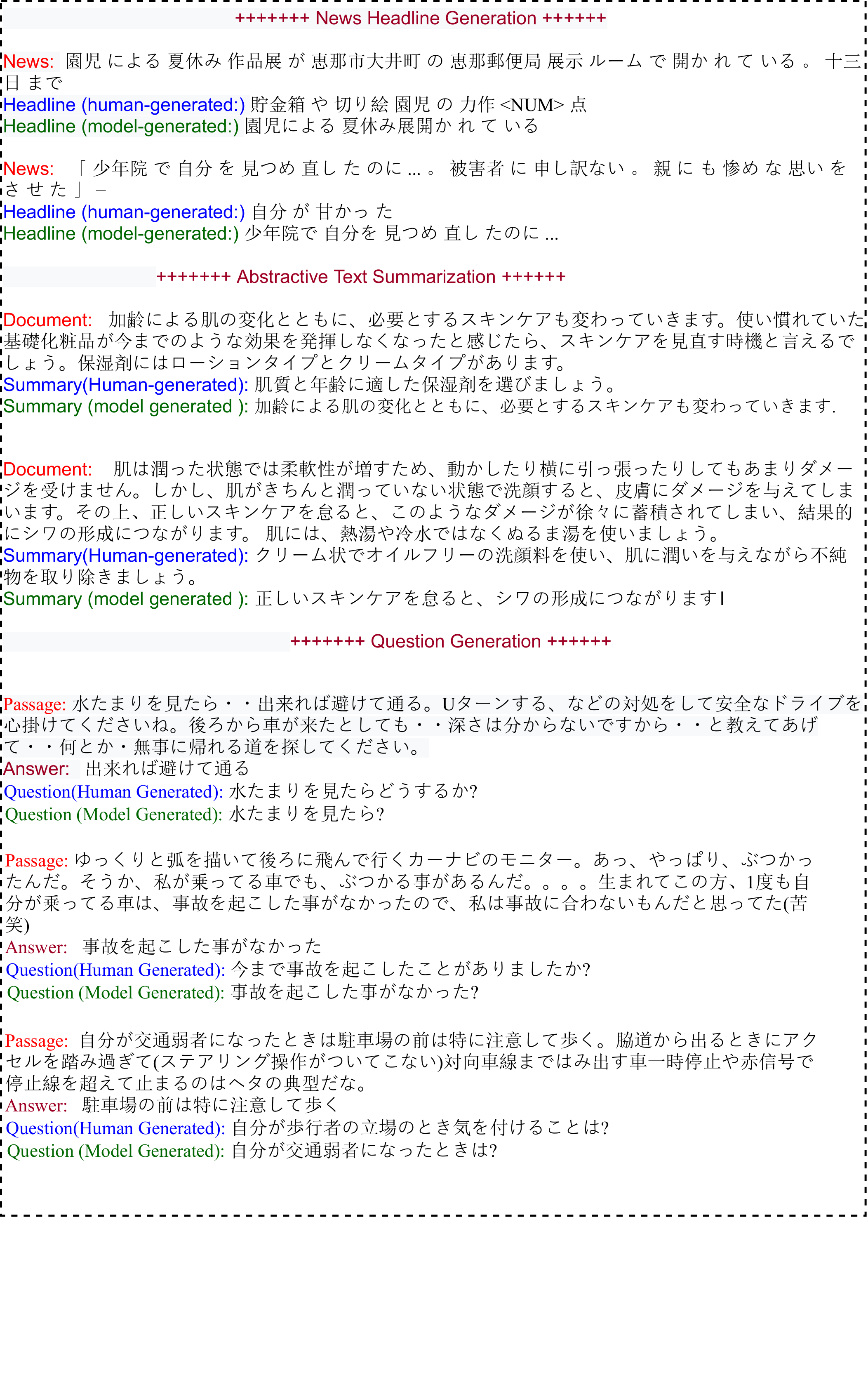}
    \label{fig:japanese_samp}
    \caption{Sample outputs for zero-shot NHG, ATS and QG in Japanese language}
\end{figure*}

\end{document}